\documentclass[11pt]{article}

\usepackage[final]{acl}

\usepackage{times}
\usepackage{latexsym}
\usepackage{amsmath}
\usepackage{amsfonts}
\usepackage{tikz}
\usepackage[T1]{fontenc}
\usepackage{makecell}
\usepackage[utf8]{inputenc}

\usepackage{microtype}

\usepackage{inconsolata}

\usepackage{graphicx}
\usepackage{booktabs}
\usepackage{mathabx}
\title{Masked Diffusion Language Models with Frequency-Informed Training}

\author{
Despoina Kosmopoulou$^{1,2}$ \quad
Efthymios Georgiou$^{3}$ \quad
Vaggelis Dorovatas$^{2}$ \quad \\[0.5ex]
\textbf{Georgios Paraskevopoulos}$^{4}$ \quad
\textbf{Alexandros Potamianos}$^{1,2}$ \\[1ex]
$^{1}$ National Technical University of Athens \\
$^{2}$ Archimedes RU, Athena RC \\
$^{3}$ University of Bern \\
$^{4}$ Institute of Language and Signal Processing, Athena RC \\
\texttt{despoinakkosmopoulou@gmail.com} \quad \texttt{efthymios.georgiou@unibe.ch} 
}

\begin{document}
\maketitle
\begin{abstract}
We present a masked diffusion language modeling framework for data-efficient training for the BabyLM 2025 Challenge. Our approach applies diffusion training objectives to language modeling under strict data constraints, incorporating frequency-informed masking that prioritizes learning from rare tokens while maintaining theoretical validity. We explore multiple noise scheduling strategies, including two-mode approaches, and investigate different noise weighting schemes within the NELBO objective. We evaluate our method on the BabyLM benchmark suite, measuring linguistic competence, world knowledge, and human-likeness. Results show performance competitive to hybrid autoregressive-masked baselines, demonstrating that diffusion-based training offers a viable alternative for data-restricted language learning. 

\end{abstract}


\section{Introduction}
By the age of 12, human children are typically exposed to fewer than 100 million words \citep{gilkerson2017mapping}. In contrast, state-of-the-art language models (LMs) \citep{touvron2023llamaopenefficientfoundation,qwen2025qwen25technicalreport} are trained on trillions of tokens. 
The BabyLM Challenge \citep{warstadt2023papersbabylmchallenge} was introduced to address this striking efficiency gap by encouraging research on more data-efficient pretraining strategies. The 2025 strict track constrains participants to train models for up to 10 epochs on a 100M-word corpus \citep{charpentier2025babylmturns3papers}.


A prominent recent approach, winning the 2024 iteration of the BabyLM Challenge, GPT-BERT, combined a Masked Language Modeling (MLM) and Next Token Prediction (NTP) objective during pretraining \citep{charpentier-samuel-2024-bert}. The MLM objective has limited learning efficiency, utilizing only ~15\% of corpus tokens per epoch \citep{devlin2019bertpretrainingdeepbidirectional}, while NTP learns from all tokens; as a result, NTP-based autoregressive (AR) generative models dominate the landscape of state-of-the-art language modeling \citep{brown2020languagemodelsfewshotlearners}. However, AR models typically use causal attention —only attending to previous tokens— which limits their bidirectional understanding and expressive ability \citep{devlin2019bertpretrainingdeepbidirectional}.

Recent advances in diffusion models have enabled their application to discrete text generation, with masked diffusion language models (MDLMs) emerging as a promising approach that combines bidirectional context modeling with generative training \citep{sahoo2024simpleeffectivemaskeddiffusion}.  
MDLMs are masked language models with ``parallel" generative capabilities, offering a compelling middle ground between the bidirectional understanding of MLMs and the generative efficiency of AR models. Unlike traditional MLM where a fixed percentage of tokens is masked at each step, MDLMs employ a diffusion process that varies masking rates across training, potentially leading to more efficient learning dynamics. 
This creates a natural curriculum where the model learns to reconstruct text under varying levels of corruption.

Recent work has shown that MDLMs can achieve competitive performance with AR models, while maintaining the bidirectional context benefits of masked models \citep{sahoo2024simpleeffectivemaskeddiffusion, shi2025simplifiedgeneralizedmaskeddiffusion}. However, diffusion models face  challenges in data-sparse settings, with their multi-step training process potentially amplifying overfitting issues—an area that remains relatively unexplored in language modeling. Specifically, MDLM effectiveness in extremely data-constrained settings remains unknown. In this work, we explore whether MDLMs trained for just 10 epochs over a 100M word corpus can match or surpass hybrid approaches like GPT-BERT.


We hypothesize that the principled diffusion training objective of MDLMs, combined with strategic masking approaches, can achieve more sample-efficient learning compared to fixed-rate MLM or purely autoregressive training. To test this hypothesis, we implement a masked diffusion language modeling framework and explore multiple noise scheduling strategies, including two-mode approaches, while investigating different noise weighting schemes within the NELBO objective. We further introduce frequency-informed masking that progressively prioritizes learning from rare tokens during the diffusion process, directing the model's attention toward more informative and challenging aspects of language while preserving the theoretical validity of the diffusion objective.


Our contributions are threefold: 1) we adapt masked diffusion language modeling for data-restricted settings, exploring multiple noise scheduling strategies including two-modes approaches and different NELBO weighting schemes, 2) we introduce a frequency-informed masking strategy that seamlessly integrates into the diffusion objective while preserving theoretical validity, and 3) we provide comprehensive evaluation on the BabyLM benchmark demonstrating that diffusion-based training achieves competitive performance with established baselines.
\section{Related Work}

\paragraph{Masked Diffusion Language Modeling:}

Inspired by continuous-time diffusion models \citep{sohldickstein2015deepunsupervisedlearningusing}, diffusion frameworks have emerged as a powerful paradigm for discrete text generation. \citet{austin2023structureddenoisingdiffusionmodels} introduced D3PM, establishing the theoretical foundation for applying diffusion to text, with concurrent work by \citet{hoogeboom2021argmaxflowsmultinomialdiffusion} and \citet{campbell2022continuoustimeframeworkdiscrete} developing discrete and continuous-time formulations.
The intersection of diffusion with masked language modeling proved particularly promising. Masked diffusion modeling formulates discrete diffusion as a Markov process with an absorbing state, where tokens replaced by MASK remain masked in subsequent steps, and the reverse process reconstructs original data from progressively corrupted representations.
\citet{sahoo2024simpleeffectivemaskeddiffusion} introduced simplified Masked Diffusion Language Models (MDLMs), unifying masked language modeling and diffusion through a simplified ELBO expression. This combines bidirectional context benefits with generative training in a unified objective. Similar simplified formulations by \citet{shi2025simplifiedgeneralizedmaskeddiffusion} and \citet{ou2025absorbingdiscretediffusionsecretly} demonstrated improved efficiency, with recent work by \citet{sahoo2025diffusionduality} bridging discrete and Gaussian diffusion for enhanced training techniques.

\paragraph{Masking Strategies for MLMs:}
Several approaches have extended BERT's 15\% random token masking \citep{devlin2019bertpretrainingdeepbidirectional} with more structured strategies. SpanBERT masks contiguous random spans rather than individual tokens and introduces a span boundary objective to predict entire masked spans \citep{joshi2020spanbertimprovingpretrainingrepresenting}, achieving substantial improvements on span selection tasks. ELECTRA replaces tokens with plausible alternatives using a generator-discriminator framework, moving beyond simple masking to token replacement detection \citep{clark2020electrapretrainingtextencoders}. RoBERTa introduces dynamic masking where different tokens are masked across training epochs, in contrast to BERT's static masking approach \citep{liu2019robertarobustlyoptimizedbert}. PMI-Masking proposes a principled approach based on Pointwise Mutual Information, jointly masking token n-grams with high collocation scores over the corpus \citep{levine2020pmimaskingprincipledmaskingcorrelated}.

\paragraph{Diffusion Models in Data-Sparse Settings:}
Diffusion models face significant challenges when applied to data-constrained scenarios. \citet{zhu2022fewshotimagegeneration} demonstrated that standard diffusion models suffer from diversity degradation in few-shot settings, leading to overfitting on limited training samples. \citet{wang2024bridgingdatagaps} identified that image-agnostic Gaussian noise creates uneven adaptation effects and proposed adversarial noise selection for more balanced transfer learning. \citet{lu2023specialistdiffusion} showed efficient adaptation through fine-tuning specific attention layers, while \citet{kulikov2023sinddm} explored single-image learning by modeling internal patch distributions. However, these findings primarily focus on vision tasks, leaving diffusion models in data-constrained LMs underexplored. 

\paragraph{Token Frequency,  Weighting and Masking:}
Frequency-based training strategies have emerged to address the imbalance of Zipfian distributions of language tokens. \citet{platanios2019competence} demonstrated that curriculum learning based on word frequency can improve sample efficiency in neural machine translation. \citet{10.1145/1553374.1553380} showed that gradually increasing task difficulty—from frequent to rare tokens—can lead to better convergence and generalization. 
Importance sampling approaches have been developed to reweight training examples based on token loss \cite{lin2024rho1}. Recent work has explored adaptive masking strategies that prioritize more salient tokens during training \cite{choi2024salience}.  However, the application of frequency-based weighting specifically to diffusion models remains underexplored, particularly in data-constrained settings where efficient learning from rare tokens becomes critical.

\section{Methodology}
\subsection{Pretraining}
\subsubsection*{Architecture} Our model architecture is a Transformer \citep{vaswani2023attentionneed}, based on the LTG-BERT model \citep{samuel2023trained100millionwords}, with the attention-gating modifications from \citep{georges-gabriel-charpentier-samuel-2023-layers}. To time-condition this model for the diffusion process, we use a timestep embedding and incorporate it with Adaptive Layer Normalization (AdaLN) modulation, following \citep{peebles2023scalable}. This approach enables the model to condition its predictions on the current masking level at timestep $t$, allowing it to adapt its behavior across different stages of the diffusion process.


\subsubsection*{Diffusion Objective} Our approach is inspired by both last year's winning GPT-BERT method and recent advances in Masked Diffusion Language Models (MDLMs) (\citealp{sahoo2024simpleeffectivemaskeddiffusion}, \citealp{shi2025simplifiedgeneralizedmaskeddiffusion}). While GPT-BERT demonstrates the effectiveness of combining encoding and generative objectives through joint training with next-token prediction and masked language modeling, MDLM success reveals that a single principled diffusion objective can achieve similar dual-purpose training. We adopt the MDLM framework to explore whether this unified approach can be effective in the data-restricted BabyLM setting.

Following the principles described by \citep{sahoo2024simpleeffectivemaskeddiffusion}, at every training step, a masking rate $1-\alpha_t$ is sampled from a distribution over $(0,1)$ for each sequence. Only masked tokens contribute to the cross-entropy loss, and the total objective is a weighted average of Masked Language Modeling (MLM) losses across different masking levels. This approach optimizes the Evidence Lower Bound (ELBO) of the diffusion process.

Specifically, in expectation, we optimize the simplified continuous-time NELBO objective from MDLM \citep{sahoo2024simpleeffectivemaskeddiffusion}:
\begin{align}
\mathcal{L} = \mathbb{E}_q\int_{t=0}^{t=1} \frac{\alpha_{t}'}{1 - \alpha_t} \sum_{\ell = 1}^{L} \log \langle \mathbf{x}_\theta^\ell(\mathbf{z}_t), \mathbf{x}^\ell \rangle \, dt
\end{align}
where $\alpha_{t}'$ denotes the time derivative of the noise schedule $\alpha_{t}$, $\mathbf{z}_t$ represents the masked sequence at time $t$, and $\mathbf{x}_\theta^\ell(\mathbf{z}_t)$ is the model's prediction for token $\ell$. This formulation provides a principled objective that naturally weights different masking levels according to the diffusion schedule, and encompasses maximum-likelihood optimization.

\subsubsection*{Frequency Informed Masking} 
We propose frequency-informed masking that assigns higher masking probabilities to rare tokens. This approach prioritizes learning from infrequent but semantically rich tokens rather than common function words. For a given sequence of tokens $t_1, \dots, t_L$ with a pre-assigned masking rate of $1-\alpha_t$, we follow a two-step process to determine the masking probability for each token. First, we rank tokens based on their global frequency, with rarer tokens receiving higher ranks. These ranks are min-max normalized to produce initial weights $\mathbf{w} \in (0,1)$. To prevent an over-emphasis on extremely rare tokens, these weights are "softened" by being raised to a power $p<1$. Our goal is to scale the weights so that they correspond to the tokens' sampling probability. 

Next, we apply conditional scaling to these weights to ensure their mean equals the target probability $1-\alpha_t$.
\begin{align}
\mathbf{w}_{\text{new}} = \begin{cases}
    \mathbf{w}^p \; \frac{1-\alpha_t}{\mu} & \text{if } \mu > 1 - \alpha_t \\
    -(1-\mathbf{w}^p)\frac{\alpha_t}{1-\mu} + 1 & \text{otherwise}
\end{cases}
\end{align}
Each token $t_i$ is then masked with a probability equal to its new weight, $w_{\text{new}_{i}}$.

This weighting scheme can be naturally extended to a form of curriculum learning \citep{10.1145/1553374.1553380} by gradually increasing the softening power $p$ from $0$ to a value $<1$ across training. This process makes the distribution of masking probabilities sharper over time, which forces the model to progressively focus on predicting rarer and more challenging tokens.

We note that frequency is only one option for the relative ranking of tokens. In our proposed framework, any masking strategy can be \textit{flexibly and seamlessly} incorporated in the Masked Diffusion LM training, or any generalized LM masking recipe.

\subsection{Evaluation}
We evaluate our framework using the BabyLM Challenge evaluation pipeline, assessing models across linguistic competence, world knowledge, human-likeness measures, and standard Natural Language Understanding (NLU) tasks. This suite tests both the quality of learned representations and their alignment with human language acquisition.
\\
\\
\noindent{\textit{Zero-Shot Evaluation.}} We evaluate our models on tasks focusing on linguistic performance and understanding, such as BLiMP \citep{warstadt2023blimpbenchmarklinguisticminimal}, Blimp Supplement \citep{warstadt2023blimpbenchmarklinguisticminimal} and a Derivational Morphology Test \citep{hofmann2024derivationalmorphologyrevealsanalogical} and a newly introduced extension \citep{weissweiler2023countingbugschatgptswugs}.
EWoK \citep{ivanova2025elementsworldknowledgeewok} tests the model's \textit{understanding} of the world, including physical concepts and causal relationships. In a similar minimal pair setting, COMPS \citep{misra-etal-2023-comps} tests inheritance of properties between hierarchical concepts. Entity Tracking \citep{kim-schuster-2023-entity} tests the model's state tracking abilities. In the zero-shot setting, the goal is for the pretrained model to assign higher likelihood for the correct sentence, from a group of sentences.
\\
\\
\noindent{\textit{Finetuning.}} The pretrained model is finetuned and evaluated on a subset of GLUE \citep{wang2019gluemultitaskbenchmarkanalysis} and SuperGLUE \citep{wang2020supergluestickierbenchmarkgeneralpurpose}, testing NLU.
\\
\\
\noindent{\textit{Human-Likeness.}} Alignment with human acquisition is of special interest when training in developmentally plausible settings. We evaluate on a Reading task using data from \citep{de_Varda2024}
and on Age of Acquisition \citep{10.1162/tacl_a_00444}. 
The derivational morphology tests \citep{hofmann2024derivationalmorphologyrevealsanalogical}, \citep{weissweiler2023countingbugschatgptswugs} provide human annotator data, and the models' higher correlation in performance with humans is assessed favorably.
\\
\\
\textit{Evaluation Backend.} In this work, we use the provided MLM backend to estimate pseudo-likelihoods of sentences \citep{Salazar_2020}. 
However, for MDLMs, this is a rather myopic view of likelihood estimation, as it only focuses on the very last denoising steps, when only one token needs to be unmasked -- on a theoretical contrast to simple MLMs, MDLMs are proper language models, able to model the whole generation process. For MDLMs, perplexity estimation can be viewed as a Monte-Carlo approximation of the diffusion denoising process \citep{sahoo2024simpleeffectivemaskeddiffusion}. Nonetheless, for the purposes of the BabyLM Challenge, the simple MLM pseudo-likelihood estimation, utilized for relatively small sentences, offers the advantage of efficient computation, sufficiently good performance, and is deterministic.
\section{Experiments}
In this section, we start by briefly describing the training and experimental setup. Afterwards, we present a series of experiments and ablations, to further explore different layers of the full framework and validate the soundness of our method.
\subsection{Training Setup}
 We use the same tokenization process, setup, optimizer and optimization hyperparameters as described in \citep{charpentier-samuel-2024-bert}, differentiating our approach in the formulation of the loss function and the masking strategy. We train all our models for 10 epochs, with a constant sequence length. The dataset in use is the BabyLM corpus.


\subsection{Experiments and Ablations}
\subsubsection*{Noise Schedules}
We train models on a linear (uniform) and a cosine masking probability schedule, and report the zero-shot results for the two configurations, evaluating them with and without time conditioning. All models are trained for 10 epochs, with a sequence length of 128.
\begin{table}[htbp]
\centering
\resizebox{0.48\textwidth}{!}{%
\begin{tabular}{lccc}
\toprule
\makecell[l]{\textbf{Noise} \\\textbf{Schedule}} & \textbf{EWoK\%} $\uparrow$& \textbf{BLiMP\%} $\uparrow$ & \textbf{BLiMP Sup.\%} $\uparrow$\\
\midrule
linear & 51.98$\pm$0.12 & 77.91$\pm$1.35 & 67.63$\pm$3.64 \\
cosine& 52.44$\pm$0.24&79.05$\pm$0.28&70.74$\pm$1.35 \\
\midrule
\makecell[l]{linear \\time cond.} & 52.16$\pm$0.51 & 77.55$\pm$0.55 & 67.23$\pm$0.98 \\
\makecell[l]{cosine\\time cond.} & 52.39$\pm$0.48 & 78.55$\pm$0.70 & 69.41$\pm$0.93 \\

\bottomrule
\end{tabular}
}
\caption{Performance comparison across different noise schedules, over 5 random seeds. Reported accuracies are field averages. Likelihoods are estimated with the standard MLM Backend. In the bottom, the results of the time conditioned evaluation are reported.}
\label{tab:performance_comparison_noise_t_cond}
\end{table}

Interestingly, the linear noise schedule, where each masking rate is equally important with the others in the loss calculation, produces relatively weak results. In the cosine schedule, masking rates are concentrated on lower values, with an expected mean of \textbf{0.36}, which is considerably smaller than the linear schedule's expected mean of \textbf{0.5}. As a result, the low-masking-rate, more fine-grained focus enables the model to perform better in the zero-shot likelihood estimation tasks, consistently.

\textit{More schedules and the importance of scaling.} In the context of finding a noise schedule that might better align with our model's learning and target tasks, we decided to experiment with unimodal and bimodal Gaussian noise schedules. This means that the distribution of $1 - \alpha_t$ is normal (or a Gaussian mixture) when $t$ is sampled uniformly. Specifically, we present a brief qualitative comparison in a small experiment of training models with a unimodal and a bimodal noise schedule with similar expected masking rates across training. ``Simple Gaussian" is a unimodal gaussian masking strategy, with masking rates coming from a $\mathcal{N}(0.3, 0.1)$ distribution. ``Bimodal Gaussian" is a mixture distribution $w_1 \mathcal{N}(\mu_1, \sigma_1^2) + (1-w_1) \mathcal{N}(\mu_2(\tau), \sigma_2^2)$ where the right mode progresses to higher values over time. In this experiment, the left mode has weight $w_1 = 0.6$, mean $\mu_1 = 0.12$, and standard deviation $\sigma_1 = 0.02$. The right mode has time-varying mean $\mu_2(\tau) = 0.4 + (0.85 - 0.4)(1 - e^{-\tau})$ and standard deviation $\sigma_2 = 0.08$, with $\tau$ representing the training progress.  

When optimizing with the full ELBO expression and including the full derivative term, $\alpha_t'$, in the calculations (p = 1.0), the zero-shot results appear underwhelming (see \autoref{tab:performance_comparison_noise_gaussian}). However, the general picture changes drastically when the derivatives are scaled down by a small power of $p$ or omitted entirely ($p=0.0$). The unimodal Gaussian schedule remains weak but improves its performance relatively. In contrast, the difference for the bimodal Gaussian case is significant. With the derivatives softened, this noise schedule allows the model to nearly reach the top-performing baseline scores. These results suggest that scaling the derivatives in the ELBO is critical for achieving better performance with certain noise schedules.

\begin{table}[htbp]
\centering
\resizebox{0.48\textwidth}{!}{%
\begin{tabular}{lccc}
\toprule
\makecell[l]{\textbf{Noise} \\\textbf{Schedule}} & \textbf{EWoK\%} $\uparrow$& \textbf{BLiMP\%} $\uparrow$ & \textbf{BLiMP Sup.\%} $\uparrow$\\
\midrule
Simple Gaussian(1.0) & 50.24 & 55.70 & 51.92 \\
Bimodal Gaussian(1.0) & 51.10 & 68.13 & 63.0\\
\midrule
\makecell[l]{Simple Gaussian(0.1)} & 50.65 & 64.34 & 59.32 \\
\makecell[l]{Bimodal Gaussian(0.1)} & 52.46 & 79.49 & 72.81\\
\midrule
\makecell[l]{Simple Gaussian(0.0)} & 50.34 & 65.34 & 58.76 \\
\makecell[l]{Bimodal Gaussian(0.0)} & 52.95 & 78.28 & 73.13 \\

\bottomrule
\end{tabular}
}
\caption{Qualitative performance comparison across different noise schedules. Reported accuracies are field averages. Likelihoods are estimated with the standard MLM Backend. \textit{($p$)} denotes the softening power $p$ for the derivative factor. Results are preliminary, run over 1 random seed.}
\label{tab:performance_comparison_noise_gaussian}
\end{table}

\subsubsection*{Frequency Informed Masking}

We compare our method's performance across two distinct configurations:
\begin{itemize}
    \item No Frequency Weighting: A baseline where tokens are masked with equal probabilities.
    \item Frequency Weighting: Our frequency-informed method is applied with a softening power of $p=0.02$, progressively (linearly) reaching this value across epochs.
\end{itemize}

We inspect the performance of these configurations on EWoK, BLiMP, and BLiMP Supplement, and report on the accuracy of the Acjective Nominalization test. All models were trained on a cosine noise schedule, with context length 128.

\begin{table}[htbp]
\centering
\small
\resizebox{0.48\textwidth}{!}{%
\begin{tabular}{lccc}
\toprule
\textbf{Config.} & \textbf{EWoK \%} $\uparrow$& \textbf{BLiMP \%} $\uparrow$ & \textbf{BLiMP Sup. \%} $\uparrow$\\
\midrule
No Freq. W. & 52.44$\pm$0.24&79.05$\pm$0.28&70.74$\pm$1.35 \\
Freq. W. & 52.63$\pm$0.36 & 78.92$\pm$0.34 & 71.77$\pm$0.86 \\
\midrule
\makecell[l]{No Freq. W.\\ time cond.} &52.39$\pm$0.48 & 78.55$\pm$0.70 & 69.41$\pm$0.93\\
\makecell[l]{Freq. W.\\ time cond.} &52.21$\pm$0.47 &78.90$\pm$0.37 &70.65$\pm$1.87\\
\bottomrule
\end{tabular}
}
\caption{Performance comparison across different token frequency weighting configurations, over 5 random seeds. The \textit{Freq. W.} configuration uses weights softened by raising the frequency distribution to power $p=0.02$ before normalization. Likelihoods are estimated with the standard MLM Backend. In the bottom, the results of the time conditioned evaluation are reported.}
\label{tab:performance_comparison_freq}
\end{table}


The frequency informed masking in general preserves or boosts performance across tasks, \textbf{improving performance on BLiMP Supplement by an absolute 1\% point} consistently. 

On the \textbf{Adjective Nominalization} test, however, we observed high variance in accuracy across random seeds. Therefore, we report a paired comparison using the same seeds. The Freq. Weight configuration evaluated with time conditioning enhances performance, \textbf{improving it by an average of 7.5 absolute percentage points}.




\subsection{Submission Model}  
\subsubsection*{Implementation}
A BPE tokenizer \citep{Gage1994} was trained with a vocabulary of 16384 tokens. The submission models have size equal to 126.6 M parameters and were trained with a fixed sequence length of 512. The batch size was set to 512, and sequences were not packed. Documents exceeding this length were divided into independent segments.  The total training duration was 10 epochs, or 7530 training steps. 

Directly following the results of our previous experiments, for the submission to the leaderboard we employed a cosine masking schedule, with $a_t = cos(\frac{\pi}{2}(1-t))$. For the frequency informed masking, we used $p=0.02$, starting from 0 at epoch 0 and linearly reaching $p$ at the last epoch. Timestep embedding dimension was set to 128.

\subsubsection*{Evaluation}
We provide\footnote{We will further update our results with the stronger bimodal gaussian schedule in our code release.} the submission's internal evaluation results, comparing them with the scores of the baseline with the maximum average score (Baseline-gpt-bert-base-mixed (mntp)). Zero-shot results were computed using the standard MLM backend without time conditioning.
\begin{table}[htbp]
\centering
\small
\resizebox{0.45\textwidth}{!}{%
\begin{tabular}{lcc}
\toprule
\textbf{Task} & \textbf{Top Baseline} & \textbf{Submission}$^\dag$ \\
\midrule
\multicolumn{3}{l}{\textbf{Linguistics}} \\
BLiMP \%                 &80.5  &76.9  \\
BLiMP Sup. \%            &73.0  &72.4  \\
\midrule
\multicolumn{3}{l}{\textbf{World Understanding}} \\
EWoK \%                  &52.4  &51.8  \\
COMPS \%                 &59.7  &56.4  \\
Entity Tracking \%       &39.9  &40.8  \\
\bottomrule
\end{tabular}%
}
\caption{Evaluation results for Linguistics and World Understanding tasks; \dag: results refer to cosine schedule}
\label{tab:sub_ling_transposed_grouped}
\end{table}

\begin{table}[htbp]
\centering
\small
\resizebox{0.45\textwidth}{!}{%
\begin{tabular}{lcc}
\toprule
\multicolumn{3}{c}{\textbf{Natural Language Understanding (Finetuning)}} \\
\midrule
\textbf{Task} & \textbf{Top Baseline} & \textbf{Submission}$^\dag$ \\
\midrule
BoolQ \%     &73.4  &72.2  \\
MNLI \%      &63.4  &63.8  \\
MRPC \%      &85.8  &88.7  \\
MultiRC \%   &69.8  &69.0  \\
QQP \%       &81.2  &79.2  \\
RTE \%       &59.0  &64.7  \\
WSC \%       &63.5  &65.4  \\
\bottomrule
\end{tabular}%
}
\caption{Evaluation results for Natural Language Understanding tasks;  \dag: results refer to cosine schedule}
\label{tab:sub_nlu_transposed}
\end{table}

\begin{table}[htbp]
\centering
\small
\resizebox{0.4\textwidth}{!}{%
\begin{tabular}{lcc}
\toprule
\textbf{Task} & \textbf{Baseline \%} $\uparrow$ & \textbf{Submission$^\dag$ \%} $\uparrow$ \\
\midrule
Reading & 6.3 & 7.4 \\
WUG Adj. N. & 41.2 & 49.6 \\
WUG Past T. & 27.1 & 15.4\\
AoA & 22.3 & -22.0 \\
\bottomrule
\end{tabular}%
}
\caption{Evaluation results for Human Likeness tasks; \dag: results refer to cosine schedule}
\label{tab:human_likeness_flipped}
\end{table}

Our model is competitive with the baseline models, particularly in the FineTuning evaluation suite. At some zero-shot evaluation tasks, the model underperforms the top-scoring baseline by relatively small margins, while it achieves better performance in Entity Tracking. On human likeness measures, the submission outperforms the top baseline in Reading and the Adjective Nominalization Test.

\section{Conclusions}
Masked Diffusion Language Models emerge as a compelling pretraining paradigm for data-constrained environments, demonstrating competitive performance against well-established baselines. Our findings reveal that the choice of masking strategy and its induced objective weighting critically determines model effectiveness. Specifically, we demonstrate that cosine noise schedules yield substantial performance gains over linear schedules, while bimodal approaches unlock even greater potential, but may require special weighting in the ELBO. Furthermore, we establish a principled framework for integrating intra-token masking strategies within the diffusion paradigm, maintaining theoretical coherence while expanding practical applicability. These results position masked diffusion as a viable path forward for efficient language model pretraining, particularly valuable when computational resources or training data are limited.

\section*{Limitations}
This work represents a conceptual integration of masked diffusion language modeling into the LTG-BERT model family, doing minimal architectural modifications. Standard implementations of masked diffusion language models often incorporate additional optimizations that can substantially impact performance; such optimizations are not explored here. Furthermore, accurately and efficiently estimating likelihoods for zero-shot tasks with short sequences using conventional diffusion approaches while maintaining low variance remains an open challenge. We hypothesize that, while the current MLM-based likelihood estimation approach captures relative trends well, it may be suboptimal, further undermining the MDLM performance.

\section*{Acknowledgments}
This work has been supported by project MIS 5154714 of the National Recovery and Resilience Plan Greece 2.0 funded by the European Union under the NextGenerationEU Program. 
We acknowledge EuroHPC JU for awarding the project ID EHPC-AI-2024A04-051 access to the EuroHPC supercomputer LEONARDO hosted by CINECA (Italy).

\bibliography{custom}

\begin{thebibliography}{44}
\providecommand{\natexlab}[1]{#1}

\bibitem[{Austin et~al.(2023)Austin, Johnson, Ho, Tarlow, and van~den Berg}]{austin2023structureddenoisingdiffusionmodels}
Jacob Austin, Daniel~D. Johnson, Jonathan Ho, Daniel Tarlow, and Rianne van~den Berg. 2023.
\newblock \href {https://arxiv.org/abs/2107.03006} {Structured denoising diffusion models in discrete state-spaces}.
\newblock \emph{Preprint}, arXiv:2107.03006.

\bibitem[{Bengio et~al.(2009)Bengio, Louradour, Collobert, and Weston}]{10.1145/1553374.1553380}
Yoshua Bengio, J\'{e}r\^{o}me Louradour, Ronan Collobert, and Jason Weston. 2009.
\newblock \href {https://doi.org/10.1145/1553374.1553380} {Curriculum learning}.
\newblock In \emph{Proceedings of the 26th Annual International Conference on Machine Learning}, ICML '09, page 41–48, New York, NY, USA. Association for Computing Machinery.

\bibitem[{Brown et~al.(2020)Brown, Mann, Ryder, Subbiah, Kaplan, Dhariwal, Neelakantan, Shyam, Sastry, Askell, Agarwal, Herbert-Voss, Krueger, Henighan, Child, Ramesh, Ziegler, Wu, Winter, Hesse, Chen, Sigler, Litwin, Gray, Chess, Clark, Berner, McCandlish, Radford, Sutskever, and Amodei}]{brown2020languagemodelsfewshotlearners}
Tom~B. Brown, Benjamin Mann, Nick Ryder, Melanie Subbiah, Jared Kaplan, Prafulla Dhariwal, Arvind Neelakantan, Pranav Shyam, Girish Sastry, Amanda Askell, Sandhini Agarwal, Ariel Herbert-Voss, Gretchen Krueger, Tom Henighan, Rewon Child, Aditya Ramesh, Daniel~M. Ziegler, Jeffrey Wu, Clemens Winter, and 12 others. 2020.
\newblock \href {https://arxiv.org/abs/2005.14165} {Language models are few-shot learners}.
\newblock \emph{Preprint}, arXiv:2005.14165.

\bibitem[{Campbell et~al.(2022)Campbell, Benton, Bortoli, Rainforth, Deligiannidis, and Doucet}]{campbell2022continuoustimeframeworkdiscrete}
Andrew Campbell, Joe Benton, Valentin~De Bortoli, Tom Rainforth, George Deligiannidis, and Arnaud Doucet. 2022.
\newblock \href {https://arxiv.org/abs/2205.14987} {A continuous time framework for discrete denoising models}.
\newblock \emph{Preprint}, arXiv:2205.14987.

\bibitem[{Chang and Bergen(2022)}]{10.1162/tacl_a_00444}
Tyler~A. Chang and Benjamin~K. Bergen. 2022.
\newblock \href {https://doi.org/10.1162/tacl_a_00444} {Word acquisition in neural language models}.
\newblock \emph{Transactions of the Association for Computational Linguistics}, 10:1--16.

\bibitem[{Charpentier et~al.(2025)Charpentier, Choshen, Cotterell, Gul, Hu, Jumelet, Linzen, Liu, Mueller, Ross, Shah, Warstadt, Wilcox, and Williams}]{charpentier2025babylmturns3papers}
Lucas Charpentier, Leshem Choshen, Ryan Cotterell, Mustafa~Omer Gul, Michael Hu, Jaap Jumelet, Tal Linzen, Jing Liu, Aaron Mueller, Candace Ross, Raj~Sanjay Shah, Alex Warstadt, Ethan Wilcox, and Adina Williams. 2025.
\newblock \href {https://arxiv.org/abs/2502.10645} {Babylm turns 3: Call for papers for the 2025 babylm workshop}.
\newblock \emph{Preprint}, arXiv:2502.10645.

\bibitem[{Charpentier and Samuel(2024)}]{charpentier-samuel-2024-bert}
Lucas Georges~Gabriel Charpentier and David Samuel. 2024.
\newblock \href {https://aclanthology.org/2024.conll-babylm.24/} {{GPT} or {BERT}: why not both?}
\newblock In \emph{The 2nd BabyLM Challenge at the 28th Conference on Computational Natural Language Learning}, pages 262--283, Miami, FL, USA. Association for Computational Linguistics.

\bibitem[{Choi et~al.(2024)Choi, Park, Yi, Cha, and Min}]{choi2024salience}
Hyesong Choi, Hyejin Park, Kwang~Moo Yi, Sungmin Cha, and Dongbo Min. 2024.
\newblock \href {https://arxiv.org/abs/2404.08327} {Salience-based adaptive masking: Revisiting token dynamics for enhanced pre-training}.
\newblock In \emph{European Conference on Computer Vision (ECCV)}, pages 343--359. Springer.

\bibitem[{Clark et~al.(2020)Clark, Luong, Le, and Manning}]{clark2020electrapretrainingtextencoders}
Kevin Clark, Minh-Thang Luong, Quoc~V. Le, and Christopher~D. Manning. 2020.
\newblock \href {https://arxiv.org/abs/2003.10555} {Electra: Pre-training text encoders as discriminators rather than generators}.
\newblock \emph{Preprint}, arXiv:2003.10555.

\bibitem[{de~Varda et~al.(2024)de~Varda, Marelli, and Amenta}]{de_Varda2024}
Andrea~Gregor de~Varda, Marco Marelli, and Simona Amenta. 2024.
\newblock \href {https://doi.org/10.3758/s13428-023-02261-8} {Cloze probability, predictability ratings, and computational estimates for 205 {E}nglish sentences, aligned with existing {EEG} and reading time data}.
\newblock \emph{Behavior Research Methods}, 56(5):5190--5213.

\bibitem[{Devlin et~al.(2019)Devlin, Chang, Lee, and Toutanova}]{devlin2019bertpretrainingdeepbidirectional}
Jacob Devlin, Ming-Wei Chang, Kenton Lee, and Kristina Toutanova. 2019.
\newblock \href {https://arxiv.org/abs/1810.04805} {Bert: Pre-training of deep bidirectional transformers for language understanding}.
\newblock \emph{Preprint}, arXiv:1810.04805.

\bibitem[{Gage(1994)}]{Gage1994}
Philip Gage. 1994.
\newblock A new algorithm for data compression.
\newblock \emph{The C Users Journal}, 12(2):23--38.

\bibitem[{Georges Gabriel~Charpentier and Samuel(2023)}]{georges-gabriel-charpentier-samuel-2023-layers}
Lucas Georges Gabriel~Charpentier and David Samuel. 2023.
\newblock \href {https://doi.org/10.18653/v1/2023.conll-babylm.20} {Not all layers are equally as important: Every layer counts {BERT}}.
\newblock In \emph{Proceedings of the BabyLM Challenge at the 27th Conference on Computational Natural Language Learning}, pages 238--252, Singapore. Association for Computational Linguistics.

\bibitem[{Gilkerson et~al.(2017)Gilkerson, Richards, Warren et~al.}]{gilkerson2017mapping}
Jill Gilkerson, Jeffrey~A. Richards, Steven~F. Warren, and 1 others. 2017.
\newblock \href {https://doi.org/10.1044/2016_AJSLP-15-0169} {Mapping the early language environment using all-day recordings and automated analysis}.
\newblock 26(2):248--265.

\bibitem[{Hofmann et~al.(2024)Hofmann, Weissweiler, Mortensen, Schütze, and Pierrehumbert}]{hofmann2024derivationalmorphologyrevealsanalogical}
Valentin Hofmann, Leonie Weissweiler, David Mortensen, Hinrich Schütze, and Janet Pierrehumbert. 2024.
\newblock \href {https://arxiv.org/abs/2411.07990} {Derivational morphology reveals analogical generalization in large language models}.
\newblock \emph{Preprint}, arXiv:2411.07990.

\bibitem[{Hoogeboom et~al.(2021)Hoogeboom, Nielsen, Jaini, Forré, and Welling}]{hoogeboom2021argmaxflowsmultinomialdiffusion}
Emiel Hoogeboom, Didrik Nielsen, Priyank Jaini, Patrick Forré, and Max Welling. 2021.
\newblock \href {https://arxiv.org/abs/2102.05379} {Argmax flows and multinomial diffusion: Learning categorical distributions}.
\newblock \emph{Preprint}, arXiv:2102.05379.

\bibitem[{Ivanova et~al.(2025)Ivanova, Sathe, Lipkin, Kumar, Radkani, Clark, Kauf, Hu, Pramod, Grand, Paulun, Ryskina, Akyürek, Wilcox, Rashid, Choshen, Levy, Fedorenko, Tenenbaum, and Andreas}]{ivanova2025elementsworldknowledgeewok}
Anna~A. Ivanova, Aalok Sathe, Benjamin Lipkin, Unnathi Kumar, Setayesh Radkani, Thomas~H. Clark, Carina Kauf, Jennifer Hu, R.~T. Pramod, Gabriel Grand, Vivian Paulun, Maria Ryskina, Ekin Akyürek, Ethan Wilcox, Nafisa Rashid, Leshem Choshen, Roger Levy, Evelina Fedorenko, Joshua Tenenbaum, and Jacob Andreas. 2025.
\newblock \href {https://arxiv.org/abs/2405.09605} {Elements of world knowledge (ewok): A cognition-inspired framework for evaluating basic world knowledge in language models}.
\newblock \emph{Preprint}, arXiv:2405.09605.

\bibitem[{Joshi et~al.(2020)Joshi, Chen, Liu, Weld, Zettlemoyer, and Levy}]{joshi2020spanbertimprovingpretrainingrepresenting}
Mandar Joshi, Danqi Chen, Yinhan Liu, Daniel~S. Weld, Luke Zettlemoyer, and Omer Levy. 2020.
\newblock \href {https://arxiv.org/abs/1907.10529} {Spanbert: Improving pre-training by representing and predicting spans}.
\newblock \emph{Preprint}, arXiv:1907.10529.

\bibitem[{Kim and Schuster(2023)}]{kim-schuster-2023-entity}
Najoung Kim and Sebastian Schuster. 2023.
\newblock \href {https://doi.org/10.18653/v1/2023.acl-long.213} {Entity tracking in language models}.
\newblock In \emph{Proceedings of the 61st Annual Meeting of the Association for Computational Linguistics (Volume 1: Long Papers)}, pages 3835--3855, Toronto, Canada. Association for Computational Linguistics.

\bibitem[{Kulikov et~al.(2023)Kulikov, Yadin, Kleiner, and Michaeli}]{kulikov2023sinddm}
Vladimir Kulikov, Shahar Yadin, Matan Kleiner, and Tomer Michaeli. 2023.
\newblock Sinddm: A single image denoising diffusion model.
\newblock In \emph{Proceedings of the 40th International Conference on Machine Learning}, pages 17920--17930. PMLR.

\bibitem[{Levine et~al.(2020)Levine, Lenz, Lieber, Abend, Leyton-Brown, Tennenholtz, and Shoham}]{levine2020pmimaskingprincipledmaskingcorrelated}
Yoav Levine, Barak Lenz, Opher Lieber, Omri Abend, Kevin Leyton-Brown, Moshe Tennenholtz, and Yoav Shoham. 2020.
\newblock \href {https://arxiv.org/abs/2010.01825} {Pmi-masking: Principled masking of correlated spans}.
\newblock \emph{Preprint}, arXiv:2010.01825.

\bibitem[{Lin et~al.(2024)Lin, Gou, Gong, Liu, Shen, Xu, Lin, Yang, Jiao, Duan, and Chen}]{lin2024rho1}
Zhenghao Lin, Zhibin Gou, Yeyun Gong, Xiao Liu, Yelong Shen, Ruochen Xu, Chen Lin, Yujiu Yang, Jian Jiao, Nan Duan, and Weizhu Chen. 2024.
\newblock \href {https://arxiv.org/abs/2404.07965} {Rho-1: Not all tokens are what you need}.
\newblock \emph{arXiv preprint arXiv:2404.07965}.

\bibitem[{Liu et~al.(2019)Liu, Ott, Goyal, Du, Joshi, Chen, Levy, Lewis, Zettlemoyer, and Stoyanov}]{liu2019robertarobustlyoptimizedbert}
Yinhan Liu, Myle Ott, Naman Goyal, Jingfei Du, Mandar Joshi, Danqi Chen, Omer Levy, Mike Lewis, Luke Zettlemoyer, and Veselin Stoyanov. 2019.
\newblock \href {https://arxiv.org/abs/1907.11692} {Roberta: A robustly optimized bert pretraining approach}.
\newblock \emph{Preprint}, arXiv:1907.11692.

\bibitem[{Lu et~al.(2023)Lu, Tunanyan, Wang, Navasardyan, Wang, and Shi}]{lu2023specialistdiffusion}
Haoming Lu, Hazarapet Tunanyan, Kai Wang, Shant Navasardyan, Zhangyang Wang, and Humphrey Shi. 2023.
\newblock Specialist diffusion: Plug-and-play sample-efficient fine-tuning of text-to-image diffusion models to learn any unseen style.
\newblock In \emph{Proceedings of the IEEE/CVF Conference on Computer Vision and Pattern Recognition (CVPR)}, pages 14267--14276.

\bibitem[{Misra et~al.(2023)Misra, Rayz, and Ettinger}]{misra-etal-2023-comps}
Kanishka Misra, Julia Rayz, and Allyson Ettinger. 2023.
\newblock \href {https://doi.org/10.18653/v1/2023.eacl-main.213} {{COMPS}: Conceptual minimal pair sentences for testing robust property knowledge and its inheritance in pre-trained language models}.
\newblock In \emph{Proceedings of the 17th Conference of the European Chapter of the Association for Computational Linguistics}, pages 2928--2949, Dubrovnik, Croatia. Association for Computational Linguistics.

\bibitem[{Ou et~al.(2025)Ou, Nie, Xue, Zhu, Sun, Li, and Li}]{ou2025absorbingdiscretediffusionsecretly}
Jingyang Ou, Shen Nie, Kaiwen Xue, Fengqi Zhu, Jiacheng Sun, Zhenguo Li, and Chongxuan Li. 2025.
\newblock \href {https://arxiv.org/abs/2406.03736} {Your absorbing discrete diffusion secretly models the conditional distributions of clean data}.
\newblock \emph{Preprint}, arXiv:2406.03736.

\bibitem[{Peebles and Xie(2023)}]{peebles2023scalable}
William Peebles and Saining Xie. 2023.
\newblock Scalable diffusion models with transformers.
\newblock \emph{Proceedings of the IEEE/CVF International Conference on Computer Vision}.

\bibitem[{Platanios et~al.(2019)Platanios, Stretcu, Neubig, Poczos, and Mitchell}]{platanios2019competence}
Emmanouil~Antonios Platanios, Otilia Stretcu, Graham Neubig, Barnabas Poczos, and Tom Mitchell. 2019.
\newblock Competence-based curriculum learning for neural machine translation.
\newblock In \emph{Proceedings of the 2019 Conference of the North American Chapter of the Association for Computational Linguistics: Human Language Technologies, Volume 1 (Long and Short Papers)}, pages 1162--1172.

\bibitem[{Qwen et~al.(2025)Qwen, :, Yang, Yang, Zhang, Hui, Zheng, Yu, Li, Liu, Huang, Wei, Lin, Yang, Tu, Zhang, Yang, Yang, Zhou, Lin, Dang, Lu, Bao, Yang, Yu, Li, Xue, Zhang, Zhu, Men, Lin, Li, Tang, Xia, Ren, Ren, Fan, Su, Zhang, Wan, Liu, Cui, Zhang, and Qiu}]{qwen2025qwen25technicalreport}
Qwen, :, An~Yang, Baosong Yang, Beichen Zhang, Binyuan Hui, Bo~Zheng, Bowen Yu, Chengyuan Li, Dayiheng Liu, Fei Huang, Haoran Wei, Huan Lin, Jian Yang, Jianhong Tu, Jianwei Zhang, Jianxin Yang, Jiaxi Yang, Jingren Zhou, and 25 others. 2025.
\newblock \href {https://arxiv.org/abs/2412.15115} {Qwen2.5 technical report}.
\newblock \emph{Preprint}, arXiv:2412.15115.

\bibitem[{Sahoo et~al.(2024)Sahoo, Arriola, Schiff, Gokaslan, Marroquin, Chiu, Rush, and Kuleshov}]{sahoo2024simpleeffectivemaskeddiffusion}
Subham~Sekhar Sahoo, Marianne Arriola, Yair Schiff, Aaron Gokaslan, Edgar Marroquin, Justin~T Chiu, Alexander Rush, and Volodymyr Kuleshov. 2024.
\newblock \href {https://arxiv.org/abs/2406.07524} {Simple and effective masked diffusion language models}.
\newblock \emph{Preprint}, arXiv:2406.07524.

\bibitem[{Sahoo et~al.(2025)Sahoo, Deschenaux, Gokaslan, Wang, Chiu, and Kuleshov}]{sahoo2025diffusionduality}
Subham~Sekhar Sahoo, Justin Deschenaux, Aaron Gokaslan, Guanghan Wang, Justin Chiu, and Volodymyr Kuleshov. 2025.
\newblock \href {https://arxiv.org/abs/2506.10892} {The diffusion duality}.
\newblock \emph{Preprint}, arXiv:2506.10892.

\bibitem[{Salazar et~al.(2020)Salazar, Liang, Nguyen, and Kirchhoff}]{Salazar_2020}
Julian Salazar, Davis Liang, Toan~Q. Nguyen, and Katrin Kirchhoff. 2020.
\newblock \href {https://doi.org/10.18653/v1/2020.acl-main.240} {Masked language model scoring}.
\newblock In \emph{Proceedings of the 58th Annual Meeting of the Association for Computational Linguistics}. Association for Computational Linguistics.

\bibitem[{Samuel et~al.(2023)Samuel, Kutuzov, Øvrelid, and Velldal}]{samuel2023trained100millionwords}
David Samuel, Andrey Kutuzov, Lilja Øvrelid, and Erik Velldal. 2023.
\newblock \href {https://arxiv.org/abs/2303.09859} {Trained on 100 million words and still in shape: Bert meets british national corpus}.
\newblock \emph{Preprint}, arXiv:2303.09859.

\bibitem[{Shi et~al.(2025)Shi, Han, Wang, Doucet, and Titsias}]{shi2025simplifiedgeneralizedmaskeddiffusion}
Jiaxin Shi, Kehang Han, Zhe Wang, Arnaud Doucet, and Michalis~K. Titsias. 2025.
\newblock \href {https://arxiv.org/abs/2406.04329} {Simplified and generalized masked diffusion for discrete data}.
\newblock \emph{Preprint}, arXiv:2406.04329.

\bibitem[{Sohl-Dickstein et~al.(2015)Sohl-Dickstein, Weiss, Maheswaranathan, and Ganguli}]{sohldickstein2015deepunsupervisedlearningusing}
Jascha Sohl-Dickstein, Eric~A. Weiss, Niru Maheswaranathan, and Surya Ganguli. 2015.
\newblock \href {https://arxiv.org/abs/1503.03585} {Deep unsupervised learning using nonequilibrium thermodynamics}.
\newblock \emph{Preprint}, arXiv:1503.03585.

\bibitem[{Touvron et~al.(2023)Touvron, Lavril, Izacard, Martinet, Lachaux, Lacroix, Rozière, Goyal, Hambro, Azhar, Rodriguez, Joulin, Grave, and Lample}]{touvron2023llamaopenefficientfoundation}
Hugo Touvron, Thibaut Lavril, Gautier Izacard, Xavier Martinet, Marie-Anne Lachaux, Timothée Lacroix, Baptiste Rozière, Naman Goyal, Eric Hambro, Faisal Azhar, Aurelien Rodriguez, Armand Joulin, Edouard Grave, and Guillaume Lample. 2023.
\newblock \href {https://arxiv.org/abs/2302.13971} {Llama: Open and efficient foundation language models}.
\newblock \emph{Preprint}, arXiv:2302.13971.

\bibitem[{Vaswani et~al.(2023)Vaswani, Shazeer, Parmar, Uszkoreit, Jones, Gomez, Kaiser, and Polosukhin}]{vaswani2023attentionneed}
Ashish Vaswani, Noam Shazeer, Niki Parmar, Jakob Uszkoreit, Llion Jones, Aidan~N. Gomez, Lukasz Kaiser, and Illia Polosukhin. 2023.
\newblock \href {https://arxiv.org/abs/1706.03762} {Attention is all you need}.
\newblock \emph{Preprint}, arXiv:1706.03762.

\bibitem[{Wang et~al.(2020)Wang, Pruksachatkun, Nangia, Singh, Michael, Hill, Levy, and Bowman}]{wang2020supergluestickierbenchmarkgeneralpurpose}
Alex Wang, Yada Pruksachatkun, Nikita Nangia, Amanpreet Singh, Julian Michael, Felix Hill, Omer Levy, and Samuel~R. Bowman. 2020.
\newblock \href {https://arxiv.org/abs/1905.00537} {Superglue: A stickier benchmark for general-purpose language understanding systems}.
\newblock \emph{Preprint}, arXiv:1905.00537.

\bibitem[{Wang et~al.(2019)Wang, Singh, Michael, Hill, Levy, and Bowman}]{wang2019gluemultitaskbenchmarkanalysis}
Alex Wang, Amanpreet Singh, Julian Michael, Felix Hill, Omer Levy, and Samuel~R. Bowman. 2019.
\newblock \href {https://arxiv.org/abs/1804.07461} {Glue: A multi-task benchmark and analysis platform for natural language understanding}.
\newblock \emph{Preprint}, arXiv:1804.07461.

\bibitem[{Wang et~al.(2024)Wang, Lin, Liu, Chen, and Xu}]{wang2024bridgingdatagaps}
Xiyu Wang, Baijiong Lin, Daochang Liu, Ying-Cong Chen, and Chang Xu. 2024.
\newblock Bridging data gaps in diffusion models with adversarial noise-based transfer learning.
\newblock In \emph{Proceedings of the 41st International Conference on Machine Learning}, pages 1--11. PMLR.

\bibitem[{Warstadt et~al.(2023{\natexlab{a}})Warstadt, Choshen, Mueller, Williams, Wilcox, and Zhuang}]{warstadt2023papersbabylmchallenge}
Alex Warstadt, Leshem Choshen, Aaron Mueller, Adina Williams, Ethan Wilcox, and Chengxu Zhuang. 2023{\natexlab{a}}.
\newblock \href {https://arxiv.org/abs/2301.11796} {Call for papers -- the babylm challenge: Sample-efficient pretraining on a developmentally plausible corpus}.
\newblock \emph{Preprint}, arXiv:2301.11796.

\bibitem[{Warstadt et~al.(2023{\natexlab{b}})Warstadt, Parrish, Liu, Mohananey, Peng, Wang, and Bowman}]{warstadt2023blimpbenchmarklinguisticminimal}
Alex Warstadt, Alicia Parrish, Haokun Liu, Anhad Mohananey, Wei Peng, Sheng-Fu Wang, and Samuel~R. Bowman. 2023{\natexlab{b}}.
\newblock \href {https://arxiv.org/abs/1912.00582} {Blimp: The benchmark of linguistic minimal pairs for english}.
\newblock \emph{Preprint}, arXiv:1912.00582.

\bibitem[{Weissweiler et~al.(2023)Weissweiler, Hofmann, Kantharuban, Cai, Dutt, Hengle, Kabra, Kulkarni, Vijayakumar, Yu, Schütze, Oflazer, and Mortensen}]{weissweiler2023countingbugschatgptswugs}
Leonie Weissweiler, Valentin Hofmann, Anjali Kantharuban, Anna Cai, Ritam Dutt, Amey Hengle, Anubha Kabra, Atharva Kulkarni, Abhishek Vijayakumar, Haofei Yu, Hinrich Schütze, Kemal Oflazer, and David~R. Mortensen. 2023.
\newblock \href {https://arxiv.org/abs/2310.15113} {Counting the bugs in chatgpt's wugs: A multilingual investigation into the morphological capabilities of a large language model}.
\newblock \emph{Preprint}, arXiv:2310.15113.

\bibitem[{Zhu et~al.(2022)Zhu, Ma, Chen, and Yuan}]{zhu2022fewshotimagegeneration}
Jingyuan Zhu, Huimin Ma, Jiansheng Chen, and Jian Yuan. 2022.
\newblock Few-shot image generation with diffusion models.
\newblock \emph{arXiv preprint arXiv:2211.03264}.

\end{thebibliography}

\end{document}